\documentclass[10pt,twocolumn,letterpaper]{article}

\usepackage[pagenumbers]{wacv} 

\usepackage{xcolor}
\definecolor{RoyalBlue}{RGB}{65, 105, 225}

\usepackage{multirow}

\definecolor{wacvblue}{rgb}{0.21,0.49,0.74}
\usepackage[pagebackref,breaklinks,colorlinks,allcolors=wacvblue]{hyperref}

\def\wacvPaperID{19} 
\def\confName{WACV}
\def\confYear{2027}

\title{
PULSE: Single-Pass Pixel-Level Uncertainty for Medical Image-to-Image Models with a Beta Output Head
\vspace{-5mm}}

\author{Noah Sizemore\\
{\normalsize Texas A\&M University-San Antonio}\\
{\tt\small nsize01@jaguar.tamu.edu
}\\
\and
Gongbo Liang\\
{\normalsize Texas A\&M University-San Antonio}\\
{\tt\small gliang@tamusa.edu}
}

\begin{document}
\maketitle


\begin{abstract}
Medical image-to-image models, such as those used for low-dose CT denoising and nucleus segmentation, produce dense predictions without indicating where those predictions may be unreliable. A clinician reviewing a denoised scan therefore cannot readily distinguish faithful anatomy from model error, which limits trust and hinders clinical deployment. Existing pixel-level uncertainty methods either require multiple forward passes, adding computational cost on clinical workstations, or use a Gaussian likelihood that assigns probability mass outside the valid $[0,1]$ intensity range. We propose \textbf{PULSE} (\textbf{P}ixel-level \textbf{U}ncertainty via \textbf{L}earned \textbf{S}ingle-pass \textbf{E}stimation), a drop-in output head that models each pixel with a Beta distribution whose support matches the normalized intensity range, yielding pixel-level uncertainty maps in a single forward pass at negligible cost. 
On low-dose CT denoising, PULSE uncertainty is rank-correlated with reconstruction error within soft tissue and bone (Spearman 0.634) and attains 90.3\% coverage at the nominal 90\% level there, and it remains rank-correlated with error with a transformer backbone (0.617). On nucleus segmentation, the same head and training objective rank errors better than a Bernoulli (sigmoid + BCE) head (AUSE 0.032 vs. 0.049). We also compare PULSE with Gaussian, truncated Gaussian, logit-normal, Kumaraswamy, multi-pass, and distribution-free heads. The Gaussian-likelihood heads in our implementation, including those inside the multi-pass baselines, use a log-variance floor that is active on CT, so the Gaussian comparisons on CT should be read with this constraint in mind.

\end{abstract}    
\section{Introduction}
\vspace{-1.mm}
Deep image-to-image (I2I) models now support many steps of medical image computing, such as low-dose CT denoising~\cite{chen2017low}, intensity standardization~\cite{liang2019ganai}, and nuclei delineation in microscopy~\cite{caicedo2019nucleus}. Their outputs increasingly inform clinical and scientific judgments, so practitioners need to know not only \emph{what} a model predicts but \emph{where} the prediction can be trusted. A radiologist reviewing a denoised low-dose CT scan, for example, must be able to identify regions where the reconstruction may be unreliable, because deep reconstruction can remove or introduce clinically relevant structures~\cite{bhadra2021hallucinations}. Yet most medical I2I models produce only deterministic point estimates, offering no pixel-level indication of confidence. This lack of transparency limits clinical trust and impedes safe deployment~\cite{wang2020inconsistent}.

\begin{figure}
    \centering
    \vspace{3.5mm}
    \includegraphics[width=1.\linewidth]{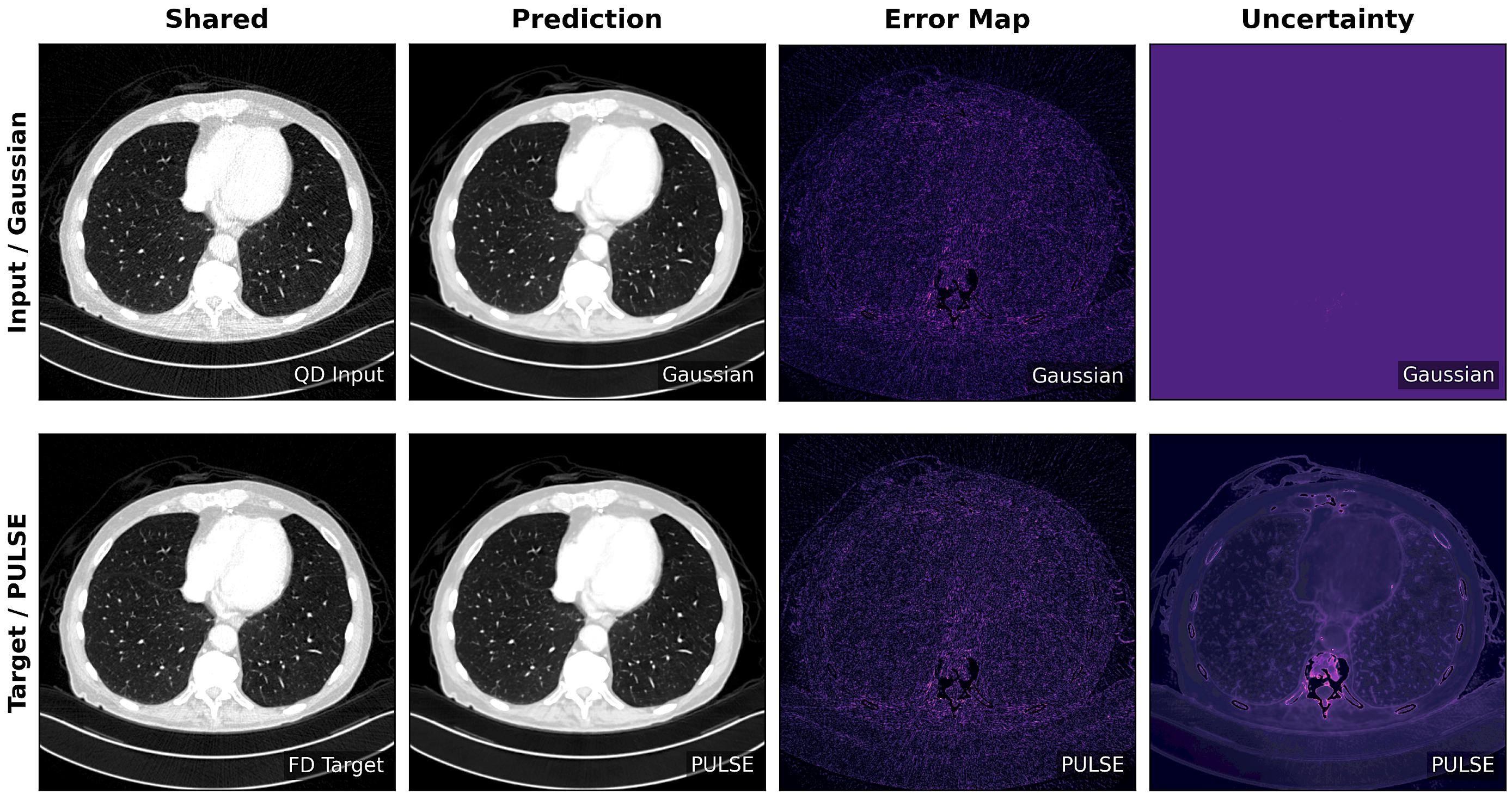}\vspace{-2.5mm}
    \caption{Gaussian and PULSE on low-dose CT denoising (Mayo LDCT~\cite{mccollough2017low}). The top row shows the quarter-dose (QD) input with the Gaussian prediction, error, and uncertainty; the bottom row shows the full-dose (FD) target with the corresponding PULSE outputs. The Gaussian uncertainty map, constrained by its log-variance floor (\textbf{Section~\ref{sec:setup}}), is nearly uniform, whereas PULSE assigns higher uncertainty to anatomical boundaries and to the high-error region near the spine. Best viewed in color.}

    \label{fig:front}
    \vspace{-1.5mm}
\end{figure}

Existing approaches to pixel-level uncertainty either require multiple forward passes at inference time~\cite{gal2016dropout, lakshminarayanan2017simple} or rely on parametric distributions whose support does not match bounded image intensities~\cite{kendall2017uncertainties}. The latter has received little attention. After windowing and normalization, CT intensities and nucleus masks lie in $[0,1]$, yet the Gaussian likelihood assigns probability mass to values outside this range. Medical images make this mismatch acute, because large regions, such as air in CT and background in microscopy images, sit at the boundary of the intensity range. We argue that this mismatch can degrade uncertainty ranking.

Building on this observation, we propose \textbf{PULSE} (\textbf{P}ixel-level \textbf{U}ncertainty via \textbf{L}earned \textbf{S}ingle-pass \textbf{E}stimation), a drop-in output head that models each pixel with a Beta distribution whose support matches the normalized intensity range. PULSE produces pixel-level uncertainty maps that indicate where a prediction can be trusted. It requires no changes to the backbone, adds negligible inference overhead, and handles continuous reconstruction and binary segmentation targets with a single training objective. We evaluate PULSE on low-dose CT denoising~\cite{mccollough2017low} and nucleus segmentation~\cite{caicedo2019nucleus}, two tasks spanning radiology and cell microscopy. In addition to Gaussian, MC Dropout, and Deep Ensemble baselines, we compare three alternative bounded distributions (truncated Gaussian, logit-normal~\cite{mead1965generalised}, and Kumaraswamy~\cite{kumaraswamy1980generalized,jones2009kumaraswamy}) and a distribution-free quantile regression head with post-hoc RCPS calibration~\cite{angelopoulos2022image}.

As shown in \textbf{Figure~\ref{fig:front}}, PULSE concentrates uncertainty where reconstruction error is high. The Gaussian map in our implementation is nearly uniform, largely because its variance floor is active on CT. Within soft tissue and bone ($\geq -500$ HU), PULSE uncertainty has a rank correlation of 0.634 with error and attains 90.3\% coverage at the nominal 90\% level. On nucleus segmentation, it achieves the lowest AUSE among all methods evaluated, including a Bernoulli head. On CT, PULSE also obtains lower AUSE than MC Dropout and Deep Ensembles (0.106 vs. 0.229 and 0.237 over all pixels) at a fraction of their inference cost, although these baselines share the Gaussian variance floor. With a U-ViT backbone, PULSE uncertainty remains rank-correlated with error (Spearman 0.617). 
Our contributions are as follows.

\begin{itemize}[leftmargin=*]

    \item We propose PULSE, a Beta output head that provides pixel-level uncertainty for medical I2I models in a single forward pass and treats continuous reconstruction and binary segmentation under one probabilistic objective.

    \item On low-dose CT denoising and nucleus segmentation, we show that PULSE yields uncertainty maps that track reconstruction error within soft tissue and bone and attain nominal coverage there.

    \item We present a controlled comparison of bounded single-pass distributions, together with multi-pass and distribution-free reference methods, on these medical tasks, suggesting that PULSE offers the most favorable balance of uncertainty ranking, inference cost, and reconstruction quality among single-pass parametric heads.
\end{itemize}
\section{Background and Related Work}
\label{sec:background}
\vspace{-1.mm}

\subsection{Aleatoric and Epistemic Uncertainty}
\vspace{-.75mm}
Predictive uncertainty in deep neural networks is commonly decomposed into \textit{aleatoric} and \textit{epistemic} components~\cite{kendall2017uncertainties}. Aleatoric uncertainty arises from inherent noise or ambiguity in the data and is irreducible regardless of model capacity or training set size~\cite{hullermeier2021aleatoric}. In medical imaging it includes acquisition noise~\cite{li2026delving,wang2019aleatoric}, such as the quantum noise that dominates low-dose CT. Epistemic uncertainty reflects the model's own ignorance in regions of the input space where training data is sparse, and is in principle reducible with additional data~\cite{kendall2017uncertainties,hullermeier2021aleatoric}. A learned output distribution captures aleatoric uncertainty directly, whereas sampling-based methods primarily capture epistemic uncertainty unless combined with a variance head.

\vspace{-.75mm}
\subsection{Sampling-Based Uncertainty Methods}
\vspace{-.75mm}
Monte Carlo (MC) dropout~\cite{gal2016dropout} approximates Bayesian inference by performing $T$ stochastic forward passes with dropout enabled at test time. Deep ensembles~\cite{lakshminarayanan2017simple} train $N$ independent models and measure predictive disagreement across members. Both are widely adopted for their simplicity, but they impose a multiplicative inference cost of $T$ or $N$ forward passes (typically $T{=}20$ and $N{=}5$), which can be prohibitive on clinical imaging workstations. Because they estimate uncertainty through model disagreement, they are most effective at capturing epistemic uncertainty~\cite{valdenegro2022deeper,durasov2021masksembles}. When aleatoric uncertainty dominates and uncertainty is estimated from model disagreement alone, ensemble members tend to converge to similar predictions, yielding overly narrow intervals~\cite{scalia2019evaluatingsu,ansari2024uncertainty}. For a fair comparison, our multi-pass baselines therefore use a Gaussian likelihood head so that their predictive variance includes an aleatoric term.

\vspace{-.75mm}
\subsection{Single-Pass Parametric Uncertainty}
\label{sec:distribution_learning}
\vspace{-.75mm}
Replacing the deterministic output with a parametric distribution enables a model to learn both a prediction and a per-pixel uncertainty in a single forward pass. Nix and Weigend~\cite{nix1994estimating} introduced this idea by predicting the mean and variance of a Gaussian, and Kendall and Gal~\cite{kendall2017uncertainties} brought the Gaussian negative log-likelihood (NLL) formulation into computer vision. Seitzer et al.~\cite{seitzer2022pitfalls} later showed that the learned variance can decouple from prediction error when the model minimizes NLL by inflating variance rather than improving the mean.

A limitation shared by all Gaussian formulations is that the distribution is defined on $(-\infty, +\infty)$, while normalized intensities are bounded in $[0,1]$. For values near 0 or 1, the Gaussian assigns probability mass to impossible values outside the valid range. Medical images make this mismatch frequent, because air in windowed CT and background in microscopy images occupy large regions at the boundary of the range. The mismatch has been noted in generative modeling. Loaiza-Ganem and Cunningham~\cite{loaiza2019continuous} introduced the continuous Bernoulli for $[0,1]$-valued pixels and compared Beta and truncated-Gaussian alternatives, and PixelCNN++~\cite{salimans2017pixelcnn++} used discretized logistic mixtures. In classical statistics, beta regression has long used a mean-precision parameterization for bounded responses~\cite{ferrari2004beta}.Beta output heads have also been used for roadway crash-risk estimation from remote-sensing imagery~\cite{elallaf2026beta}. However, no prior work has systematically evaluated how distributional support and shape affect \emph{pixel-level uncertainty ranking} in dense image-to-image models.

There are other single-pass approaches. Evidential deep learning~\cite{amini2020deep,sensoy2018evidential} places a Normal-Inverse-Gamma prior over Gaussian likelihood parameters to decompose uncertainty in one pass, but it inherits the unbounded Gaussian support. Mixture density networks~\cite{bishop1994mixture} and quantile regression~\cite{koenker2001quantile} offer more flexible or distribution-free alternatives, and Angelopoulos et al.~\cite{angelopoulos2022image} combine quantile regression with risk-controlling prediction sets (RCPS) to obtain distribution-free per-pixel intervals for image-to-image regression. Laplace likelihoods are bounded on one side and lack the shape flexibility of the Beta family. We compare bounded parametric families (Beta, truncated Gaussian, logit-normal, and Kumaraswamy) against Gaussian and distribution-free baselines under identical conditions.

\subsection{Uncertainty in Medical Image Reconstruction and Analysis}
\vspace{-1.25mm}
Deep reconstruction and enhancement models can introduce structures absent from the measured data, a failure mode formalized for tomographic imaging as hallucination~\cite{bhadra2021hallucinations}, and the uncertainty of medical image generation models has been analyzed through their learned feature space~\cite{qu2022beware}. Pixel-level uncertainty has been proposed as a safeguard. Edupuganti et al.~\cite{edupuganti2021uncertainty} derive pixel variance maps for MRI reconstruction by sampling from a variational autoencoder, and Tanno et al.~\cite{tanno2021uncertainty} combine a heteroscedastic noise model with approximate Bayesian inference for neuroimage enhancement, showing that predictive uncertainty correlates with reconstruction error. These approaches rely on Gaussian likelihoods or repeated sampling. We instead study a single-pass head whose support matches bounded medical intensities, and evaluate it on both reconstruction (CT denoising) and analysis (nucleus segmentation).

\subsection{Uncertainty Evaluation and Calibration}
\label{sec:uncertainty_evaluation}
\vspace{-1.25mm}
For uncertainty estimates to be useful in practice, they must be \emph{calibrated}, meaning that predicted uncertainty should faithfully reflect the likelihood of prediction error~\cite{guo2017calibration, levi2022evaluating}. For medical imaging classifiers, calibration has been improved through trainable regularizers~\cite{liang2020improved} and probabilistic embeddings~\cite{han2024improving}. Miscalibrated uncertainty can mislead users, with overconfident models suppressing warranted caution and underconfident models rendering the uncertainty map too noisy to be actionable~\cite{lopez2025uncertainty, jiang2011calibrating}.

Reliability diagrams~\cite{kuleshov2018accurate, liang2020improved} plot predicted uncertainty against empirical error across bins, and the Expected Calibration Error (ECE)~\cite{guo2017calibration, naeini2015obtaining} summarizes the deviation as a scalar. Prediction Interval Coverage Probability at 90\% (PICP@90) measures whether predicted 90\% intervals contain the true value at the expected rate~\cite{pearce2018high}. Proper scoring rules such as NLL jointly evaluate calibration and sharpness but are distribution-specific~\cite{gneiting2007strictly}. The Area Under the Sparsification Error curve (AUSE)~\cite{ilg2018uncertainty} instead progressively removes pixels in order of predicted uncertainty and measures how quickly the remaining error decreases. Because AUSE does not depend on the functional form of the predicted distribution, it enables fair comparison across methods and serves as our primary metric. In medical images, all-pixel metrics can also be dominated by clinically irrelevant regions such as background air, so we additionally report region-restricted metrics where appropriate.
\begin{figure}
    \centering
    \includegraphics[width=.92\linewidth]{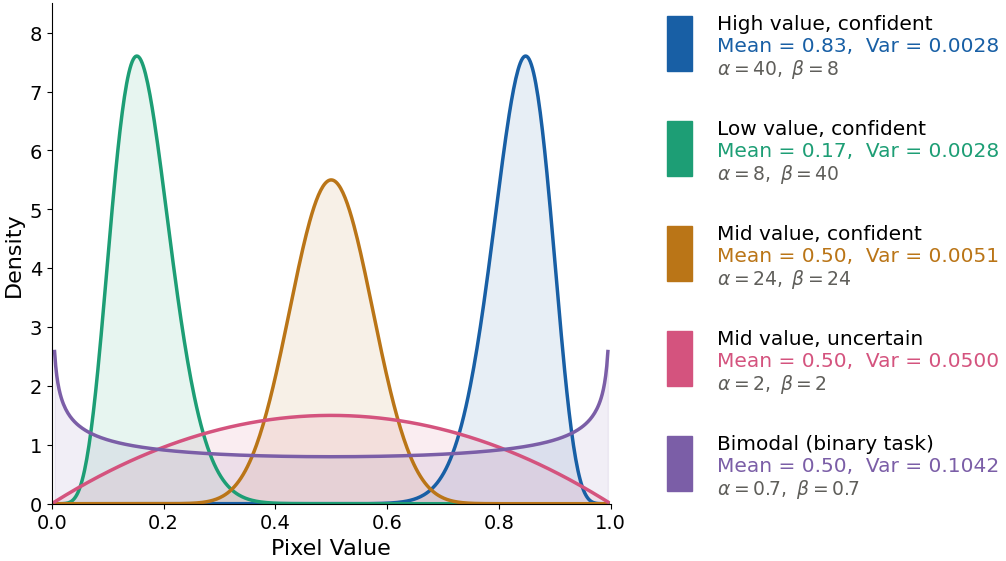}\vspace{-3.mm}
    \caption{Example Beta densities. Varying $\alpha$ and $\beta$ produces confident (tight), uncertain (broad), skewed, and bimodal distributions, all bounded within $[0,1]$ by construction.}\vspace{-3.5mm}
    \label{fig:beta_shapes}
\end{figure}

\section{Method}
\label{sec:method}
\vspace{-1.75mm}
We present PULSE, a lightweight probabilistic output head that produces pixel-level uncertainty maps for medical image-to-image models. We formalize the problem, then describe the head, its training objective, and the extraction of uncertainty maps at inference.

\vspace{-.75mm}
\subsection{Problem Formulation}
\label{sec:problem-setup}
\vspace{-1.mm}
Let $\mathbf{x} \in \mathbb{R}^{H \times W \times C_{\text{in}}}$ denote an input image and $\mathbf{y} \in [0,1]^{H \times W \times C_{\text{out}}}$ the corresponding target. Medical targets are naturally bounded after standard preprocessing. CT intensities are clipped to a fixed Hounsfield unit (HU) window and rescaled to $[0,1]$, and nucleus segmentation masks take values in $\{0,1\}$. A standard model $f_\theta$ produces a deterministic estimate $\hat{\mathbf{y}} = f_\theta(\mathbf{x})$ with no measure of confidence. We instead predict, for each pixel $i$, a distribution $p_\theta(y_i \mid \mathbf{x})$ whose mean serves as the prediction and whose spread indicates where that prediction can be trusted. The distribution should satisfy two properties: (1) its support should match the valid range $[0,1]$, and (2) it should be obtainable in a single forward pass without changes to the backbone.

\vspace{-.75mm}
\subsection{PULSE Output Head}
\vspace{-1.mm}
PULSE models each pixel as an independent Beta distribution, $\text{Beta}(\alpha, \beta)$, which is defined on exactly the interval $[0,1]$ with probability density function
%
\begin{equation}
  p(y \mid \alpha, \beta)
  = \frac{y^{\alpha-1}(1-y)^{\beta-1}}{B(\alpha,\beta)},
  \quad y \in (0,1),
\label{eq:beta_pdf}
\end{equation}
where $B(\alpha,\beta)$ is the Beta function and $\alpha, \beta > 0$ are shape parameters. As illustrated in \textbf{Figure~\ref{fig:beta_shapes}}, varying $\alpha$ and $\beta$ yields tight, broad, skewed, and bimodal densities within the valid range. Skewed densities suit intensities near the boundaries of the range, such as air in CT or background in microscopy images, where a Gaussian must either place probability mass outside $[0,1]$ or inflate its variance. The concentration $\alpha + \beta$ also controls confidence independently of the mean $\frac{\alpha}{\alpha+\beta}$, so a mid-range prediction can be reported either as a confident estimate (a sharp peak) or as an uncertain one (a broad density). This distinction is invisible in the point prediction, yet it is the information a clinician needs to decide where to look more closely.

\begin{figure}[!tb]
    \centering
    \includegraphics[width=1.\linewidth]{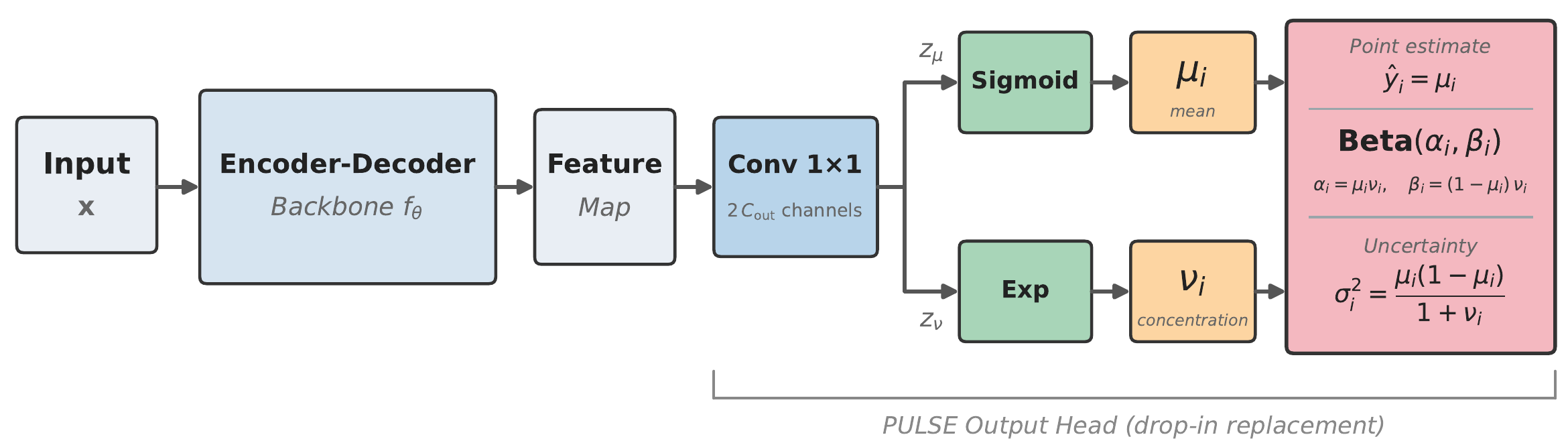}\vspace{-2.5mm}
    \caption{PULSE architecture. The backbone is unchanged. A $1 \times 1$ convolution on the final feature map predicts a mean $\mu$ (sigmoid) and a concentration $\nu$ (exponential) for each pixel, which define $\text{Beta}(\mu\nu, (1-\mu)\nu)$. The Beta mean is the prediction and the Beta variance is the uncertainty map.}
    \vspace{-1.5mm}
    \label{fig:architecture}
\end{figure}

As shown in \textbf{Figure~\ref{fig:architecture}}, PULSE leaves the backbone $f_\theta$ unchanged and adds a $1 \times 1$ convolution to its final feature map, producing two raw outputs $z_{\mu,i}$ and $z_{\nu,i}$ per pixel and output channel. Following the mean-precision parameterization of beta regression~\cite{ferrari2004beta}, these are mapped to a mean $\mu_i$ and a concentration $\nu_i = \alpha_i + \beta_i$,
%
\begin{equation}
\begin{aligned}
  \mu_i &= \text{sigmoid}(z_{\mu,i}), &\quad \nu_i &= \exp(z_{\nu,i}), \\
  \alpha_i &= \mu_i \nu_i, &\quad \beta_i &= (1 - \mu_i)\,\nu_i.
\end{aligned}
\label{eq:param}
\end{equation}
For numerical stability, $\mu_i$ is clipped to $[10^{-6}, 1 - 10^{-6}]$ and $z_{\nu,i}$ to $[-7, 7]$, which caps the concentration at $e^7$. The prediction and its uncertainty follow directly as the Beta mean and variance,
%
\begin{equation}
  \mathbb{E}[y_i] = \mu_i, \quad
  \sigma_i^2 = \frac{\mu_i (1 - \mu_i)}{1 + \nu_i}.
\label{eq:mean_var}
\end{equation}
\textbf{Equation~\ref{eq:mean_var}} makes the decoupling explicit, where $\mu_i$ sets the predicted value and $\nu_i$ scales confidence around it. Predicting $\alpha_i$ and $\beta_i$ directly is an alternative that we compare in our ablations (\textbf{Section~\ref{sec:ablations}}). Because PULSE modifies only the output layer and loss, it is in principle compatible with any encoder-decoder backbone. We evaluate it on a convolutional U-Net and transformer-based U-ViT.

\subsection{Training Objective}
\label{sec:training}
\vspace{-.75mm}
PULSE is trained by minimizing the negative log-likelihood (NLL) of the observed target pixels under the predicted Beta distributions. For a pixel with target $y_i$ and parameters $\alpha_i, \beta_i$ from \textbf{Equation~\ref{eq:param}}, the loss is
%
\begin{equation}
\begin{split}
  \mathcal{L}_i =
    & -(\alpha_i - 1)\log y_i
      - (\beta_i - 1)\log(1 - y_i) \\
    & + \log B(\alpha_i, \beta_i).
\end{split}
\label{eq:nll}
\end{equation}
The total loss is averaged over all pixels in a mini-batch. Because the logarithms are undefined at the boundaries, targets are clamped to $[\epsilon, 1-\epsilon]$ with $\epsilon = 10^{-6}$, which matters for CT, where voxels below the HU window saturate at 0. For nucleus segmentation, where targets lie in $\{0, 1\}$, the same objective applies unchanged after clamping. At fixed mean, the Beta distribution approaches a Bernoulli as its concentration decreases, so a single objective handles continuous reconstruction and binary segmentation.

\subsection{Uncertainty Map Extraction}
\vspace{-.75mm}
At inference, the variance $\sigma_i^2$ from \textbf{Equation~\ref{eq:mean_var}} serves as the pixel-level uncertainty map. Low variance marks regions where the prediction can be relied upon, and high variance flags regions that warrant closer review. This map is used for all ranking-based metrics and reliability diagrams described in \textbf{Section~\ref{sec:uncertainty_evaluation}}.

For prediction interval evaluation, we compute the 90\% central interval from the Beta quantile function $F^{-1}$,
%
\begin{equation}
  [l_i, u_i] = \left[
    F^{-1}(0.05;\, \alpha_i, \beta_i),\;
    F^{-1}(0.95;\, \alpha_i, \beta_i)
  \right].
\label{eq:pi}
\end{equation}
A target $y_i$ is covered if $l_i \leq y_i \leq u_i$. Unlike Gaussian intervals, which can extend below 0 or above 1, Beta intervals remain within the valid range by construction.
\section{Experiments}
\label{sec:experiments}
\vspace{-1mm}

\begin{table*}[t]
\centering
\caption{Main results on low-dose CT denoising (Mayo LDCT) and nucleus segmentation (DSB 2018) with a U-Net backbone. Best in \textbf{bold}, second-best \underline{underlined}. $\dagger$ indicates multi-pass inference. CT uncertainty metrics use all pixels; see \textbf{Table~\ref{tab:tissue}} for tissue-masked results. All NLL values include the Gaussian normalizing constant. The Gaussian, MC Dropout, and Deep Ensemble heads share a log-variance floor that binds on 99.7\% of CT test pixels for the Gaussian head, so their CT uncertainty metrics reflect this floor. Multi-pass and binary-target coverage are not reported (\textbf{Section~\ref{sec:setup}}).}
\label{tab:main}
\vspace{-1.5mm}
\setlength{\tabcolsep}{7.75pt}
\begin{tabular}{l ccccc ccc c}
\toprule
& \multicolumn{5}{c}{\textbf{CT denoising}} & \multicolumn{3}{c}{\textbf{Nucleus segmentation}} & \\
\cmidrule(lr){2-6} \cmidrule(lr){7-9}
\textbf{Method} & PSNR $\uparrow$ & SSIM $\uparrow$ & AUSE $\downarrow$ & NLL $\downarrow$ & PICP@90 & Dice $\uparrow$ & IoU $\uparrow$ & AUSE $\downarrow$ & Cost \\
\midrule
Deterministic           & \textbf{35.49} & 0.869 & ---   & ---     & ---   & \underline{0.919} & \underline{0.850} & ---   & 1$\times$ \\
Gaussian                & \underline{35.45} & \textbf{0.876} & \underline{0.460} & $-$2.49 & 0.995 & 0.913 & 0.839 & \underline{0.046} & 1$\times$ \\
sigmoid + BCE           & ---   & ---   & ---   & ---     & ---   & \textbf{0.920} & \textbf{0.851} & 0.049 & 1$\times$ \\
MC Dropout$^\dagger$    & 35.31 & \textbf{0.876} & 0.229 & $-$2.49 & ---   & 0.899 & 0.817 & 0.048 & 20$\times$ \\
Deep Ensemble$^\dagger$ & 35.28 & 0.870 & 0.237 & $-$2.49 & ---   & 0.911 & 0.837 & 0.047 & 5$\times$ \\
\textbf{PULSE}          & 35.27 & \underline{0.875} & \textbf{0.106} & \textbf{$-$4.72} & 0.712 & 0.904 & 0.825 & \textbf{0.032} & 1$\times$ \\
\bottomrule
\end{tabular}
\vspace{-1.5mm}
\end{table*}

\begin{table}[t]
\centering
\caption{Gaussian and PULSE on clinically focused measures. Left: CT metrics within soft tissue and bone ($\geq -500$ HU). Right: mean Spearman correlation between uncertainty and error. Gaussian CT values reflect its log-variance floor, which binds on 99.3\% of pixels in this region (\textbf{Section~\ref{sec:setup}}).}
\label{tab:tissue}
\vspace{-1.5mm}
\setlength{\tabcolsep}{1.5pt}
\begin{tabular}{l ccc cc}
\toprule
& \multicolumn{3}{c}{\textbf{CT, tissue-masked}} & \multicolumn{2}{c}{\textbf{Spearman} $\uparrow$} \\
\cmidrule(lr){2-4} \cmidrule(lr){5-6}
\textbf{Method} & PICP@90 & MPIW $\downarrow$ & AUSE $\downarrow$ & CT & Nucleus \\
\midrule
Gaussian       & 0.989 & 0.098 & 0.541 & 0.040 & 0.640 \\
\textbf{PULSE} & 0.903 & \textbf{0.061} & \textbf{0.294} & \textbf{0.634} & \textbf{0.993} \\
\bottomrule
\end{tabular}
\vspace{-.75mm}
\end{table}

\subsection{Experimental Setup}
\label{sec:setup}
\vspace{-.75mm}

\paragraph{Tasks and datasets.} For \emph{low-dose CT denoising}, we use the 2016 NIH-AAPM-Mayo Clinic Low-Dose CT dataset~\cite{mccollough2017low} (Mayo LDCT), mapping quarter-dose 1\,mm slices to their full-dose counterparts. Intensities are clipped to $[-1000, 400]$ HU, following RED-CNN~\cite{chen2017low}, and rescaled to $[0,1]$. We use a fixed patient-level split with eight patients for training, one for validation (L333), and one for testing (L506). For \emph{nucleus segmentation}, we use the 2018 Data Science Bowl (DSB) dataset~\cite{caicedo2019nucleus}, merging per-nucleus masks into one binary mask per image, with a seeded 80/10/10 image-level split. We treat nucleus segmentation as binary dense prediction, where the Beta mean gives the per-pixel foreground probability. Both tasks are trained on $192 \times 192$ patches. To test whether the gains of bounded heads depend on boundary-heavy targets, 
we also train the single-pass heads on four natural-image datasets with few pixels near 0 or 1, namely grayscale-to-color colorization on CIFAR-10~\cite{krizhevsky2009learning} and STL-10~\cite{coates2011stl10}, and $4\times$ super-resolution on DIV2K~\cite{agustsson2017ntire} and BSD100~\cite{martin2001database}.

\vspace{-3.75mm}
\paragraph{Backbone and training.} All methods share a U-Net~\cite{ronneberger2015u} with batch normalization~\cite{ioffe2015batch}, and only the output head and loss differ, isolating the output distribution's effect. We also test a U-ViT~\cite{bao2023all} on CT. Models are trained with Adam~\cite{kingma2014adam} (learning rate $2\times10^{-4}$, batch size 16, or 128 for colorization) and cosine annealing.

\vspace{-3.75mm}
\paragraph{Baselines.} We compare PULSE with a deterministic U-Net trained with L1 loss (a prediction-quality reference), a Gaussian head~\cite{kendall2017uncertainties} trained with Gaussian NLL, and three alternative bounded heads (truncated Gaussian, logit-normal~\cite{mead1965generalised}, and Kumaraswamy~\cite{kumaraswamy1980generalized,jones2009kumaraswamy}). We also include MC Dropout ($T{=}20$, $p{=}0.2$) and Deep Ensemble ($N{=}5$), both built on Gaussian heads so that their total predictive variance combines the mean member variance (aleatoric) with the variance of member means (epistemic). On CT we further evaluate quantile regression with post-hoc RCPS calibration~\cite{angelopoulos2022image}, and on nucleus segmentation a sigmoid + BCE (Bernoulli) head.

\vspace{-3.75mm}
\paragraph{Implementation note.} For numerical stability, all heads bound their raw outputs with hard clamps, which have zero gradient beyond the bound. The Gaussian-likelihood heads, including those inside MC Dropout and Deep Ensemble, clamp the log-variance at $-7$, a minimum standard deviation of 0.030 in normalized units. Typical CT reconstruction errors are smaller than this floor, and on the test patient 99.3\% of pixels within soft tissue and bone have a predicted log-variance at the floor. Gaussian uncertainty maps, coverage, interval widths, and NLL on CT therefore reflect this floor rather than a learned variance, and comparisons with Gaussian-likelihood heads on CT are comparisons against this implementation. On nucleus segmentation, the floor binds on 77.1\% of test pixels, almost entirely background (85.4\% of background versus 6.4\% of nucleus pixels), where errors are near zero, and above the floor the Gaussian standard deviation spans a 22-fold range. PULSE's concentration cap binds on 23.3\% of CT pixels within soft tissue and bone (median concentration 709). Within this region, PULSE's variance has a rank correlation of $-0.14$ with $\mu(1-\mu)$, so its uncertainty map is not a function of the predicted intensity alone.

\begin{figure}[t]
  \centering
  \includegraphics[width=.95\columnwidth]{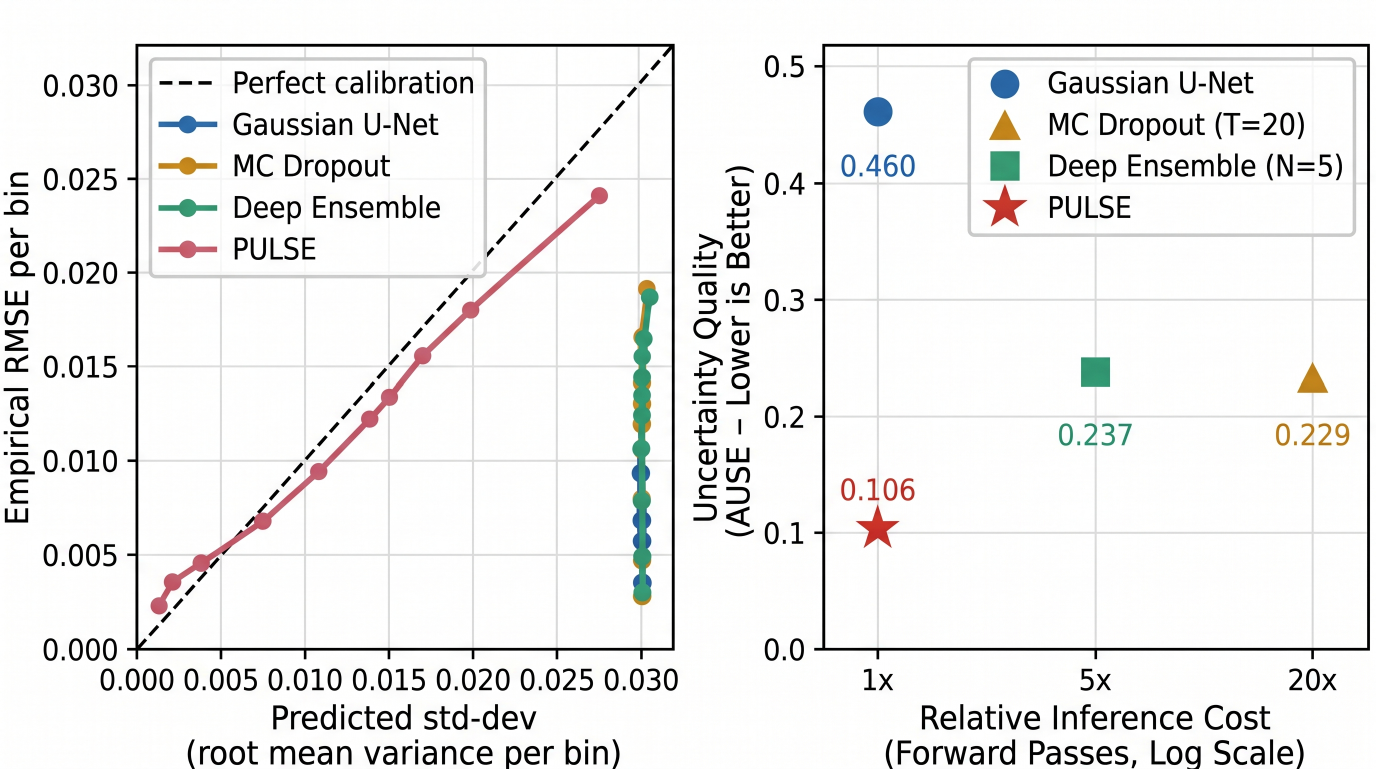}\vspace{-1.5mm}
  \caption{CT denoising. \textbf{Left:} Reliability diagram. The Gaussian, MC Dropout, and Deep Ensemble curves overlap at a nearly constant predicted standard deviation, whereas PULSE tracks empirical error. \textbf{Right:} All-pixel AUSE versus relative inference cost.}
  \label{fig:reliability}
  \vspace{-2.5mm}
\end{figure}

\begin{figure*}[t]
  \centering
  \includegraphics[width=.75\linewidth]{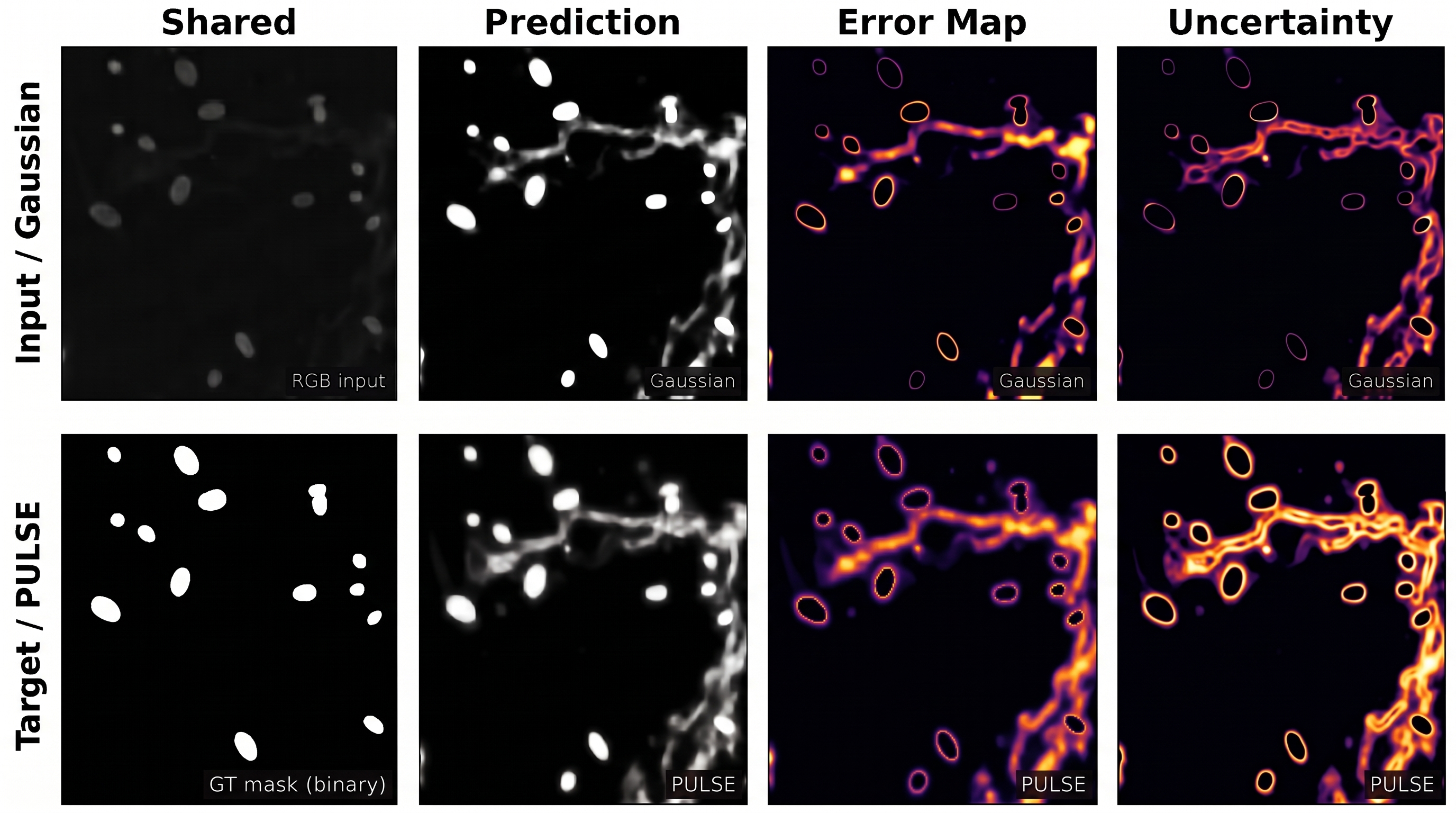}\vspace{-1.5mm}
  \caption{Nucleus segmentation (DSB 2018). The top row shows the input with the Gaussian prediction, error, and uncertainty; the bottom row shows the ground-truth mask with the PULSE outputs. The PULSE prediction is the Beta mean, a per-pixel foreground probability. Both models segment a filamentous structure absent from the ground truth, and PULSE assigns it high uncertainty. Gaussian uncertainty is faint on several nucleus boundaries where its error is high, whereas PULSE uncertainty follows its errors. Best viewed in color.}
  \label{fig:nucleus}
\end{figure*}

\vspace{-3.75mm}
\paragraph{Evaluation.} We report PSNR and SSIM~\cite{wang2004image} for CT and Dice and IoU for nucleus segmentation. AUSE~\cite{ilg2018uncertainty} is our primary uncertainty metric, complemented by NLL, PICP@90, MPIW, and the Spearman correlation between uncertainty and error. About 23\% of CT test pixels lie at the $-1000$ HU normalization floor (air outside the patient), so all-pixel CT metrics are dominated by clinically irrelevant background. We therefore also report metrics within soft tissue and bone ($\geq -500$ HU, 38.7\% of pixels). We do not report interval coverage for binary targets, where central intervals around clamped $\{0,1\}$ targets are not meaningful, or for the multi-pass baselines, which we compare on ranking quality only.

\vspace{-3.75mm}
\paragraph{Data and ethics.} Both datasets are publicly available and de-identified. Mayo LDCT was released for the 2016 AAPM Low Dose CT Grand Challenge~\cite{mccollough2017low}, and DSB 2018 consists of publicly released microscopy images~\cite{caicedo2019nucleus}. No new data were collected, so institutional review board approval was not required. We used both datasets in accordance with their data-use terms.

\subsection{Low-Dose CT Denoising}
\label{sec:ct}
\vspace{-1.25mm}

\paragraph{Uncertainty ranking.} PULSE reduces all-pixel AUSE from 0.460 to 0.106 relative to the Gaussian head at identical cost (\textbf{Table~\ref{tab:main}}). Within soft tissue and bone, its AUSE is 0.294 versus 0.541 (\textbf{Table~\ref{tab:tissue}}), and its uncertainty has a rank correlation of 0.634 with error, compared with 0.040 for the Gaussian. Both heads produce visually indistinguishable reconstructions (\textbf{Figure~\ref{fig:front}}), but the Gaussian uncertainty map is nearly uniform, whereas the PULSE map concentrates on anatomical boundaries and on the high-error region near the spine. The near-uniform Gaussian map and the Gaussian values above largely reflect its variance floor (\textbf{Section~\ref{sec:setup}}).

\vspace{-3.75mm}
\paragraph{Multi-pass methods.} 
MC Dropout and Deep Ensemble reduce AUSE relative to the Gaussian head but remain behind PULSE (0.229 and 0.237 vs. 0.106) at 5$\times$ to 20$\times$ the inference cost. Their member heads share the Gaussian variance floor, so their aleatoric term is close to constant and their ranking comes mainly from member disagreement. As \textbf{Figure~\ref{fig:reliability}} shows, their predicted standard deviation stays near the floor. Whether PULSE retains its advantage over multi-pass baselines with unconstrained variance heads remains to be tested.

\vspace{-3.75mm}
\paragraph{Coverage and calibration.} PULSE's all-pixel PICP@90 of 0.712 is depressed by the normalization floor, where the model correctly predicts near-zero variance and even quantization noise falls outside the collapsed interval. Within soft tissue and bone, PULSE attains 90.3\% coverage with intervals 38\% narrower than the Gaussian's (MPIW 0.061 vs.\ 0.098), whereas the Gaussian over-covers (98.9\%) with intervals whose width is close to that implied by its variance floor (about 0.099). In the reliability diagram (\textbf{Figure~\ref{fig:reliability}}), PULSE's predicted standard deviation follows empirical error across a wide range, while the Gaussian's predicted standard deviation stays at its floor regardless of error.

\vspace{-3.75mm}
\paragraph{Likelihood and distribution-free comparison.} PULSE also attains lower NLL than the Gaussian head ($-4.72$ vs.\ $-2.49$), 
although NLL values are not strictly comparable across distributional families~\cite{gneiting2007strictly}, and the Gaussian NLL on CT is limited by its variance floor. 
The quantile + RCPS baseline obtains a slightly lower AUSE (0.098), and RCPS raises its coverage to nominal without changing the ranking. It requires a held-out calibration set and provides intervals rather than a full predictive density, whereas PULSE yields an analytic density in a single pass with no post-hoc stage.

\vspace{-3.75mm}
\paragraph{Prediction quality.} PULSE stays within 0.22\,dB PSNR of the deterministic U-Net with comparable SSIM (0.875 vs.\ 0.869), so the probabilistic head costs little reconstruction accuracy.

\begin{table*}[t]
\centering
\caption{AUSE of bounded single-pass heads (lower is better). Best in \textbf{bold}, second-best \underline{underlined}. Natural-image datasets are included to test whether the gains depend on boundary-heavy targets. The Gaussian Mayo CT entry reflects its log-variance floor (\textbf{Section~\ref{sec:setup}}).}

\label{tab:bounded}
\vspace{-2.5mm}
\setlength{\tabcolsep}{12.5pt}
\begin{tabular}{l cc cccc}
\toprule
& \multicolumn{2}{c}{\textbf{Medical images}} & \multicolumn{4}{c}{\textbf{Natural images}} \\
\cmidrule(lr){2-3} \cmidrule(lr){4-7}
\textbf{Method} & Mayo CT & Nucleus & CIFAR-10 & STL-10 & DIV2K & BSD100 \\
\midrule
Gaussian           & 0.460 & 0.046 & 0.333 & 0.241 & \underline{0.022} & \textbf{0.124} \\
Truncated Gaussian & 0.185 & 0.235 & 0.344 & \underline{0.159} & \underline{0.022} & 0.130 \\
Logit-normal       & \textbf{0.074} & 0.052 & \underline{0.317} & \textbf{0.133} & 0.036 & 0.136 \\
Kumaraswamy        & 0.155 & \underline{0.034} & 0.959 & 0.371 & 0.027 & 0.159 \\
\textbf{PULSE}     & \underline{0.106} & \textbf{0.032} & \textbf{0.302} & 0.224 & \textbf{0.021} & \underline{0.126} \\
\bottomrule
\end{tabular}
\end{table*}

\begin{figure*}[t]
  \centering
  \includegraphics[width=.95\linewidth]{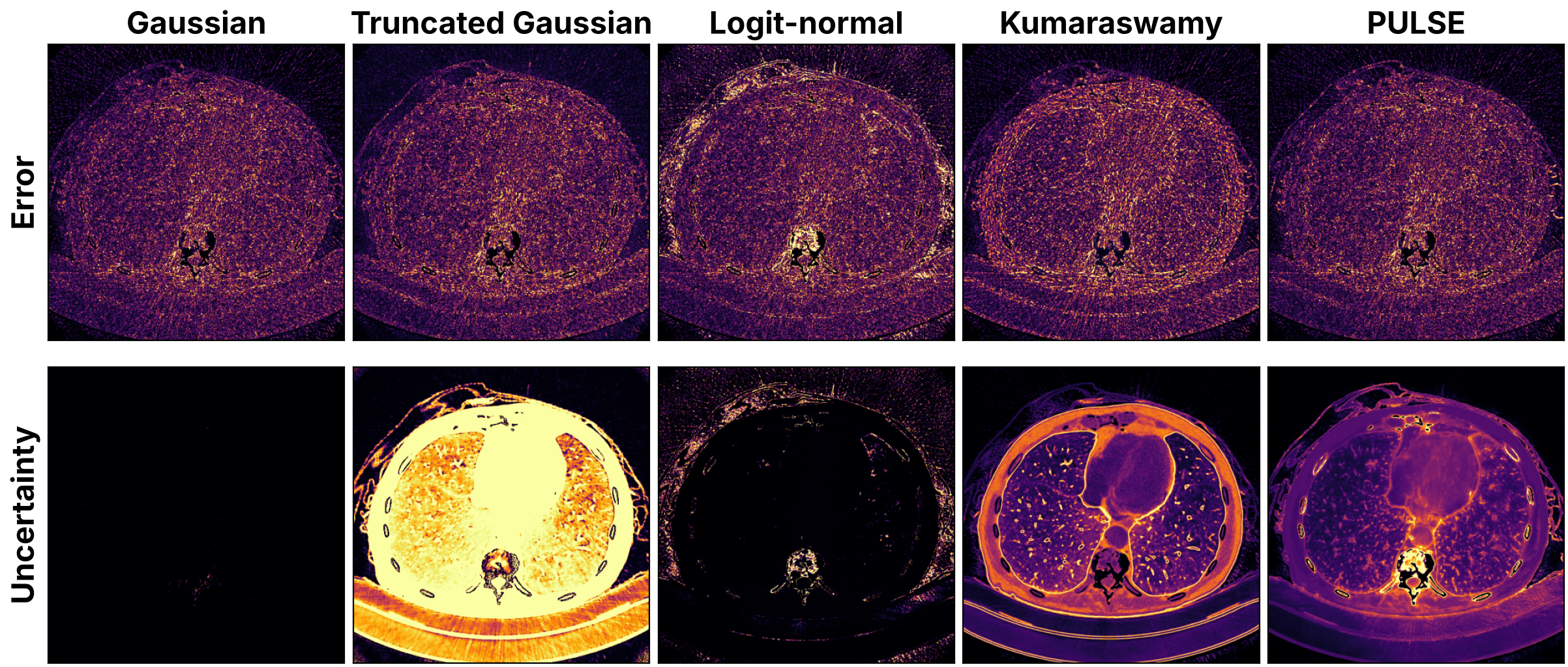}\vspace{-1.5mm}
  \caption{Bounded single-pass heads on a CT test slice. Top: absolute error of each head. Bottom: predicted uncertainty, each map shown on its own color scale because magnitudes differ by more than an order of magnitude; per-map scaling preserves the pixel ranking that AUSE evaluates. The Gaussian map, constrained by its log-variance floor (\textbf{Section~\ref{sec:setup}}), is nearly uniform, the truncated Gaussian saturates across the body, and the logit-normal map is sparse with speckle outside the body. Kumaraswamy also produces a structured map but with higher pixel error, whereas PULSE concentrates its highest uncertainty at the spine, where error is largest. Best viewed in color.}
  \label{fig:ct_heads}
  \vspace{-1mm}
\end{figure*}

\subsection{Nucleus Segmentation}
\label{sec:nucleus}
\vspace{-.75mm}

PULSE obtains the lowest AUSE among all methods (0.032; \textbf{Table~\ref{tab:main}}), including the correctly specified sigmoid + BCE head (0.049) and both multi-pass baselines, while using the same head and objective as for CT. Its uncertainty is nearly perfectly rank-correlated with error (0.993 vs.\ 0.640 for the Gaussian; \textbf{Table~\ref{tab:tissue}}). As \textbf{Figure~\ref{fig:nucleus}} shows, PULSE assigns high uncertainty to a non-nuclear structure that both models falsely segment, and its uncertainty follows boundary errors that the Gaussian uncertainty misses. The Bernoulli head attains slightly higher Dice (0.920 vs.\ 0.904), as expected for a likelihood matched to binary targets. These results show that a single Beta head handles both continuous reconstruction and binary segmentation.


\subsection{Comparison of Bounded Distributions}
\label{sec:bounded}
\vspace{-.75mm}
PULSE's improvement over the Gaussian could stem from bounded support, shape flexibility, or the mean-concentration parameterization. To separate these factors, we replace the output head with three alternatives that also respect $[0,1]$, trained under identical conditions (\textbf{Table~\ref{tab:bounded}}). On CT, every bounded head improves substantially over the Gaussian, which is consistent with bounded support mattering on intensities concentrated near the boundary, although part of this gap may reflect the Gaussian variance floor. Bounded support alone is not sufficient, however. The truncated Gaussian degrades sharply on nucleus segmentation (0.235), and the logit-normal, which ranks first on CT (0.074), falls behind the Gaussian on nucleus segmentation (0.052) and costs prediction quality (33.46 vs.\ 35.27\,dB PSNR on CT and 0.870 vs.\ 0.904 Dice). PULSE is the only head that ranks in the top two on both tasks while preserving prediction quality. \textbf{Figure~\ref{fig:ct_heads}} illustrates these differences on a CT slice. The truncated Gaussian saturates across the body and the logit-normal map is sparse and noisy. Kumaraswamy also yields a structured map, but with 0.7\,dB lower PSNR than PULSE (34.56 vs.\ 35.27\,dB) and higher AUSE (0.155 vs.\ 0.106), whereas PULSE concentrates its uncertainty where error is largest.

On the natural-image datasets, where few pixels lie near 0 or 1, the advantage of bounded support shrinks. Bounded heads improve most on CT, where about 23\% of pixels lie at the normalization floor, and give little or no benefit on super-resolution, where the Gaussian is best or tied (0.124 on BSD100, 0.022 on DIV2K). Across all six datasets, PULSE ranks first or second on five.

\begin{table*}[t]
\centering
\caption{CT denoising with a U-ViT backbone. Best in \textbf{bold}, second-best \underline{underlined}. $\dagger$ indicates multi-pass inference. Spearman is the mean rank correlation between uncertainty and error. The Gaussian, MC Dropout, and Deep Ensemble heads use the same log-variance floor as with the U-Net (\textbf{Section~\ref{sec:setup}}).}
\label{tab:uvit}
\vspace{-1.5mm}
\setlength{\tabcolsep}{12.5pt}
\begin{tabular}{l cc cc c}
\toprule
\textbf{Method} & PSNR $\uparrow$ & SSIM $\uparrow$ & AUSE $\downarrow$ & Spearman $\uparrow$ & Cost \\
\midrule
Deterministic             & \underline{34.91} & 0.863 & ---   & ---   & 1$\times$ \\
Gaussian                  & \textbf{35.03} & \textbf{0.871} & 0.405 & 0.028 & 1$\times$ \\
MC Dropout$^\dagger$      & 34.98 & 0.870 & 0.155 & 0.545 & 20$\times$ \\
Deep Ensemble$^\dagger$   & 35.00 & 0.870 & 0.188 & 0.506 & 5$\times$ \\
\textbf{PULSE}            & 33.52 & 0.847 & \textbf{0.097} & \textbf{0.617} & 1$\times$ \\
\textbf{PULSE, direct $(\alpha,\beta)$} & 34.71 & \underline{0.867} & \underline{0.128} & ---   & 1$\times$ \\
\bottomrule
\end{tabular}
\end{table*}

\subsection{Backbone and Ablations}
\label{sec:ablations}
\vspace{-0.75mm}

\paragraph{Transformer backbone.} With a U-ViT backbone on CT denoising, PULSE again obtains the lowest AUSE (0.097, vs.\ 0.405 for the Gaussian, 0.155 for MC Dropout, and 0.188 for Deep Ensemble) and the highest rank correlation with error (0.617; \textbf{Table~\ref{tab:uvit}}). Its PSNR is 1.4\,dB below the deterministic U-ViT (33.52 vs.\ 34.91), a larger gap than with the U-Net. The direct $(\alpha, \beta)$ variant comes within 0.2\,dB of the deterministic model while still outperforming all baselines on AUSE (0.128), indicating that the quality gap is not inherent to the Beta head. The U-ViT Gaussian heads use the same log-variance floor, and their near-zero rank correlation with error (0.028) suggests that it binds here as well, although we did not measure this directly.

\begin{figure}[!t]
    \centering
    \includegraphics[width=1.\linewidth]{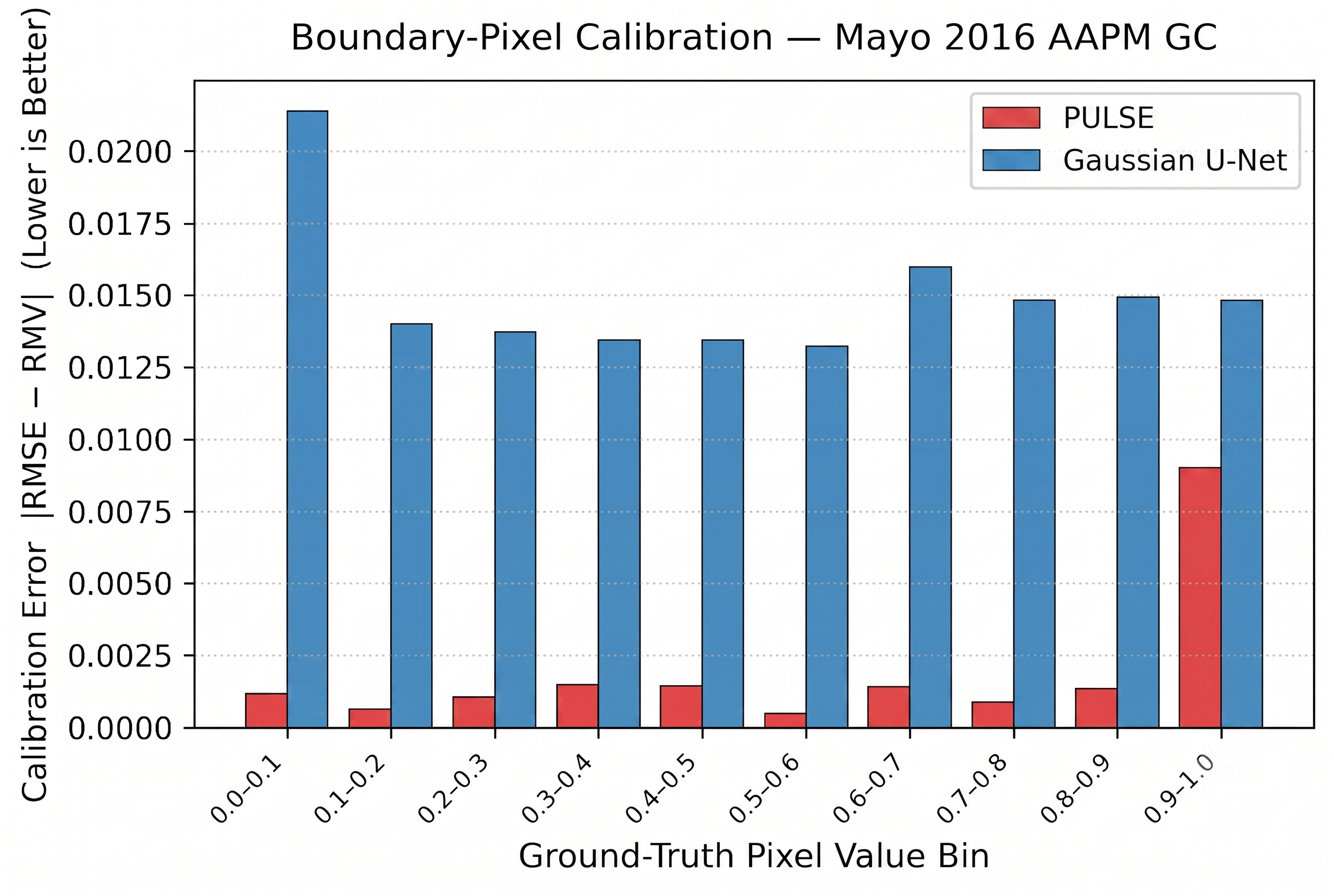}\vspace{-4.mm}
    \caption{Boundary pixel analysis on CT denoising. Per-bin discrepancy $|\text{RMSE}-\text{RMV}|$ for Gaussian and PULSE.}
    \vspace{-2.75mm}
    \label{fig:boundary}
\end{figure}

\vspace{-3.75mm}
\paragraph{Boundary pixel analysis.} To examine behavior near the extremes of the intensity range, we bin CT test pixels by ground-truth value into ten equal-width bins and compute the per-bin discrepancy $|\text{RMSE} - \text{RMV}|$ (\textbf{Figure~\ref{fig:boundary}}). PULSE has lower discrepancy than the Gaussian in every bin. The Gaussian discrepancy peaks in the $[0.0, 0.1]$ bin, where errors are near zero while its standard deviation is held at the floor, and remains elevated in interior bins, in line with its nearly constant variance at the floor. 

\vspace{-3.75mm}
\paragraph{Additional ablations.} The direct $(\alpha, \beta)$ and mean-concentration $(\mu, \nu)$ parameterizations perform comparably on both tasks (AUSE 0.109 vs.\ 0.110 on CT and 0.034 vs.\ 0.032 on nucleus segmentation). We adopt $(\mu, \nu)$ because it separates the predicted value from the confidence and improved optimization stability. On CIFAR-10 colorization, results were identical for boundary clamping values $\epsilon$ from $10^{-8}$ to $10^{-4}$, so $\epsilon$ requires no tuning. PULSE adds less than 0.05\,ms per image over the deterministic baseline (9.68 vs.\ 9.63\,ms on an NVIDIA A30 GPU), whereas MC Dropout and Deep Ensemble require about 20$\times$ and 5$\times$ the computation.
\section{Conclusion}
\vspace{-1.mm}
We introduced PULSE, a drop-in output head that provides pixel-level uncertainty maps for medical image-to-image models in a single forward pass. By modeling each pixel with a Beta distribution whose support matches the normalized intensity range, PULSE indicates where a reconstruction or segmentation can be trusted. On low-dose CT denoising, its uncertainty tracked error within soft tissue and bone (Spearman 0.634) and attained 90.3\% coverage at the nominal 90\% level there. On nucleus segmentation, the same head and objective achieved the lowest AUSE among all methods evaluated, including a Bernoulli head. PULSE also ranked errors better than our multi-pass baselines at 5$\times$ to 20$\times$ the inference cost, and it was the only bounded head among those tested to rank in the top two on both tasks.

Such trust maps could support clinical workflows, for example by directing a reader's attention to regions of a denoised scan that warrant closer review or by prioritizing cases for a second read. Because PULSE adds negligible inference cost, it can run wherever the underlying model runs, including on clinical imaging workstations. These uses motivate the work but remain to be validated in a reader study.

\vspace{-3.75mm}
\paragraph{Limitations and future work.}
The Gaussian-likelihood heads, including those inside MC Dropout and Deep Ensembles, used a log-variance floor that is active on CT, and PULSE used a cap on concentration. The Gaussian and multi-pass comparisons on CT are therefore not against unconstrained heads, and repeating them with relaxed bounds and variance-stabilized objectives such as $\beta$-NLL~\cite{seitzer2022pitfalls} is the most important next step. The CT results rest on a single dataset with one test patient, and tissue-masked metrics were computed only for the Gaussian and PULSE heads. All-pixel coverage on CT (71.2\%) is depressed by background at the normalization floor, which shows that all-pixel metrics can mislead when large portions of an image are clinically irrelevant; evaluation protocols that account for region relevance deserve further study. We did not conduct a clinician reader study, so the clinical value of the uncertainty maps is motivated rather than demonstrated. PULSE models aleatoric uncertainty with a single network, so it does not capture epistemic uncertainty and is not designed to flag out-of-distribution inputs such as scans from unseen scanners or protocols~\cite{wang2020inconsistent, liang2019ganai}. Combining PULSE with ensembles or MC Dropout would capture both components, and comparing such variants against multi-pass Gaussian baselines at matched cost would isolate the effect of bounded support when epistemic uncertainty is also modeled. PULSE also treats pixels independently and does not capture spatial correlation in uncertainty. With a U-ViT backbone, the default $(\mu, \nu)$ parameterization incurred a larger PSNR gap than with a U-Net (1.4\,dB), although the direct $(\alpha, \beta)$ variant largely closed it. Finally, because PULSE modifies only the output layer, it can in principle be attached to pretrained biomedical foundation models, as probabilistic adaptation has been applied to vision-language models for radiograph retrieval~\cite{elallaf2026medprobclip}, which we leave to future work.

\section*{Acknowledgments}
Portions of this research utilized high-performance computing resources provided by the High Performance Computing Research Center at Texas A\&M University-San Antonio, led by Dr. Izzat Alsmadi. We gratefully acknowledge his support in facilitating access to these resources.

{
    \small
    \bibliographystyle{ieeetr}    
    \bibliography{main, liang}
}

\end{document}